\documentclass[conference]{IEEEtran}
\makeatletter
\long\def\@makecaption#1#2{\ifx\@captype\@IEEEtablestring%
\footnotesize\begin{center}{\normalfont\footnotesize #1}\\
{\normalfont\footnotesize\scshape #2}\end{center}%
\@IEEEtablecaptionsepspace
\else
\@IEEEfigurecaptionsepspace
\setbox\@tempboxa\hbox{\normalfont\footnotesize {#1.}~~ #2}%
\ifdim \wd\@tempboxa >\hsize%
\setbox\@tempboxa\hbox{\normalfont\footnotesize {#1.}~~ }%
\parbox[t]{\hsize}{\normalfont\footnotesize \noindent\unhbox\@tempboxa#2}%
\else
\hbox to\hsize{\normalfont\footnotesize\hfil\box\@tempboxa\hfil}\fi\fi}
\usepackage[compatibility=false]{caption}
\DeclareCaptionFont{quackfont}{\fontfamily{ptm}\fontsize{7.5pt}{9pt}\selectfont}
\usepackage[labelformat=simple,font=quackfont]{subcaption}

\makeatother
\usepackage{lipsum}
\makeatletter
\let\MYcaption\@makecaption
\usepackage{cite}
\usepackage{graphicx}
\usepackage{subcaption}
\usepackage{float}
\usepackage{refstyle}
\usepackage{multicol, blindtext}
\usepackage{amsmath,amssymb,amsfonts}
\usepackage{adjustbox}

\usepackage[font=scriptsize]{caption}
\usepackage[font=footnotesize]{caption}
\usepackage{stfloats}
\usepackage{lmodern,bm}                
\usepackage[T1]{sansmath}
\usepackage{xcolor}
\usepackage{algorithm, algorithmic}
\usepackage{enumitem}
\usepackage{dutchcal}

\usepackage{array}
\usepackage{multirow}

\begin{document}

\title{Skeleton Split Strategies for Spatial Temporal Graph Convolution Networks}

\author{\IEEEauthorblockN{Motasem S. Alsawadi}
\IEEEauthorblockA{Electronic and Electrical Engineering Department\\
Roberts Building, University College London\\
Torrington Place, London \\
WC1E 7JE\\
motasem.alsawadi.18@ucl.ac.uk}
\and
\IEEEauthorblockN{Miguel Rio}
\IEEEauthorblockA{Electronic and Electrical Engineering Department\\
Roberts Building, University College London\\
Torrington Place, London \\
WC1E 7JE\\
miguel.rio@ucl.ac.uk}
}
\maketitle

\begin{abstract}
A skeleton representation of the human body has been proven to be effective for this task. The skeletons are presented in graphs form-like. However, the topology of a graph is not structured like Euclidean-based data. Therefore, a new set of methods to perform the convolution operation upon the skeleton graph is presented. Our proposal is based upon the ST-GCN framework proposed by Yan \textit{et al.} \cite{yan2018spatial}. In this study, we present an improved set of label mapping methods for the ST-GCN framework. We introduce three split processes (full distance split, connection split, and index split) as an alternative approach for the convolution operation. To evaluate the performance, the experiments presented in this study have been trained using two benchmark datasets: NTU-RGB+D and Kinetics. Our results indicate that all of our split processes outperform the previous partition strategies and are more stable during training without using the edge importance weighting additional training parameter. Therefore, our proposal can provide a more realistic solution for real-time applications centered on daily living recognition systems activities for indoor environments.  
\end{abstract}

\IEEEpeerreviewmaketitle

\section{Introduction}
\IEEEPARstart{A}{ction} recognition (AR) has been recognized as an activity in which individuals' behavior can be observed. Assembling profiles of regular activities such as activities of daily living (ADL) can support identifying trends in the data during critical events. These include actions that might compromise a person's life. For that reason, human action recognition has become an active research area. Generally, human activity is characterized by different recipes. These include optical flows, appearance, and body skeletons. \cite{kinoshita2006tracking,bobick1997movement,yan2018spatial}. Amidst these recipes, dynamic human skeletons (DHS) usually carry vital information that encompasses other modalities. One of the main benefits of this approach is that it minimizes the need for wearing sensors. Therefore, to collect the data, surveillance cameras can be mounted on the ceiling or walls of the environment of interest; ensuring an efficient indoor monitoring system \cite{foroughi2008intelligent}. However, DHS modeling has not yet been fully explored.\par 

A performed action is typically described by a time series of the 2D or 3D coordinates of human joint positions \cite{li2019actional,yan2018spatial}. Furthermore, action is recognized by examining the motion patterns. A skeleton representation of the human body has been proven to be effective for this task. It provides a robust solution to noise, and it is considered to be a computational and storage-efficient solution\cite{li2019actional}. Additionally, it provides a background-free data representation to the classification algorithms. This allows the algorithms to focus only on the human body pattern recognition without being concerned about the surrounding environment of the performed action scenarios. This work aims to develop a unique and efficient approach for modeling the DHS for human action recognition.\par 

\subsection{Open Pose}
There are multiple sources of camera-based skeleton data. Recently, Cao \textit{et al.} \cite{cao2019openpose} released the open-source library \emph{OpenPose} which allows real-time skeleton-based human detection. Their algorithm outputs the skeleton graph represented as an array with the 2D and the 3D coordinates. They are 18 tuples with values (X, Y, C) for 2D and (X, Y, Z, C) for 3D; where C is the confidence score of the detected joint, X, Y and Z represent the coordinates on the X-axis, Y-axis and the Z-axis of the video frame, respectively. 

\subsection{Spatial Temporal Graph Neural Network}
New techniques have been proposed recently to exploit the connections between the joints of a skeleton. Among these, Convolutional Neural Networks (CNNs) are used to address human action modeling tasks due to their ability to automatically capture the patterns contained in the spatial configuration of the joints and their temporal dynamics \cite{tu2018skeleton}. However, the skeletons are presented in graphs form-like, making it difficult to use conventional CNNs to model the dynamics of human actions. Thanks to the recent evolution of Graph Convolutional Neural Networks (GCNNs), it is possible to analyze the non-structured data in an end-to-end manner. These techniques generalize CNNs to the graphs structures\cite{si2019attention}.\par

\begin{figure}[!t]
    \centering
    \includegraphics[width=0.25\textwidth]{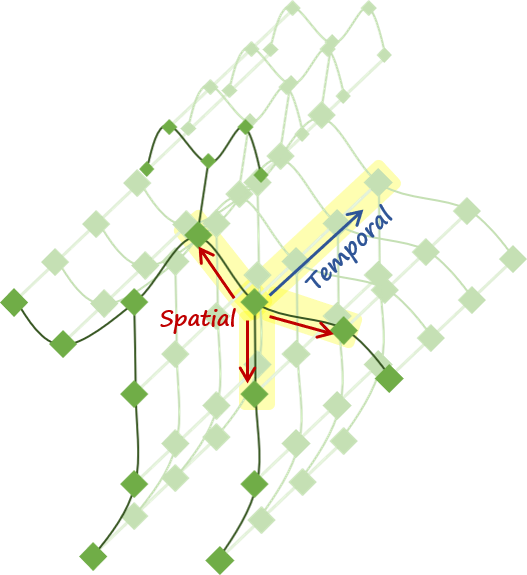}
    \caption{Spatiotemporal graph representation of a skeleton.}
    \label{fig:stgcn1}
\end{figure}

In order to achieve an accurate ADL recognition, the temporal dimension has to be considered. An action can be considered as a time-dependent pattern of a set of joints in motion\cite{li2019actional}. Given the advantages of GCNs mentioned previously, numerous approaches for skeleton-based action recognition using this architecture have been proposed. The first GCN-based solution for action recognition using skeleton data was presented by Yan \emph{et al.} \cite{yan2018spatial}. They considered both spatial and temporal dimensions of skeleton joints movements at the modeling stage. This approach is called the Spatiotemporal Graph Convolutional Network (ST-GCN) model. In the ST-GCN model, every joint has a set of edges for the spatial and temporal dimensions independently, as it is illustrated in Fig. \ref{fig:stgcn1}.  Suppose a given sequence of frames with skeleton joints coordinates; then the spatial edges connect each joint with its neighborhood per frame. On the other hand, temporal boundaries connect each joint with another joint corresponding to the exact location from a consecutive frame. Meaning that the temporal edge set represents the joint trajectory over time \cite{yan2018spatial}. However, the topology of the graph is not implicitly structured like Euclidean-based data. For instance, most of the nodes have different numbers of neighbors. Therefore, multiple strategies for applying the convolution operation upon skeleton joints have been proposed.\par  

\begin{figure}[!t]
    \centering
    \includegraphics[width=0.35\textwidth]{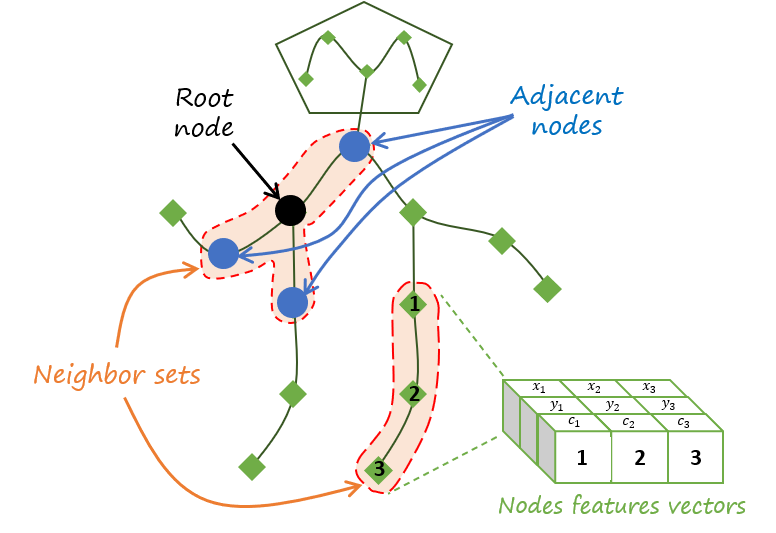}
    \caption{Skeleton components.}
    \label{fig:skeleton_components}
\end{figure}

In their work, Yan \textit{et al.} \cite{yan2018spatial} presented multiple solutions to perform the convolution operation over the skeleton graph. They first divided the skeleton graph into a fixed subset of nodes (the skeleton joints) they called \emph{neighbor sets}. Every neighbor set has a central node (the \emph{root node}) and its adjacent nodes. Subsequently, it is performed a partitioning of the neighbor set into a fixed number of $K$ subsets, where a numeric label (which we call \emph{priority}) is assigned to each of them. Formally, each adjacent node $u_{ti}$ in a neighbor set $B(u_{tj})$ of a root node $u_{tj}$ is \emph{mapped to a label} $l_{ti}$. On the other hand, each filter of the CNN has a $K$ number of subsets of values. Therefore, each subset of values of a filter performs the convolution operation process upon the feature vector of its corresponding node. Given that the skeleton data has been obtained using the Open Pose toolbox \cite{cao2019openpose}, each feature vector consists of the 2D coordinates of the joints, including a value of confidence $C$. These ideas are illustrated in Fig.\ref{fig:skeleton_components}.

\begin{figure}[!t]
    \centering
    \includegraphics[width=0.5\textwidth]{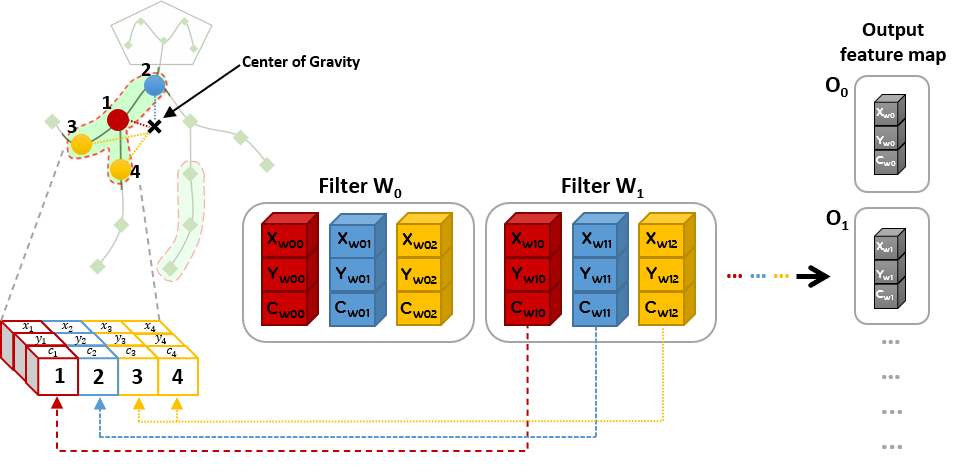}
    \caption{Spatial configuration partitioning.}
    \label{fig:spatial_partitioning}
\end{figure}

\subsubsection*{Spatial configuration partitioning strategy}
In this strategy, the partitioning for the label mapping is performed according to the distance of each node in the neighbor set with respect to the center of gravity \(cg\) of the skeleton graph. According to \cite{yan2018spatial}, each neighbour set is divided into three (filter size K = 3). Therefore, each kernel has three subsets of values; one for the root node, one for the joints closer to \(cg\) and another one for the joints located farther with respect to \(cg\). As it can be seen in Fig. \ref{fig:spatial_partitioning}, each filter with three subsets of values is applied to the node feature vectors in order to create the output feature map. In this technique, the filter size $K=3$, and the mapping are defined by the following \cite{yan2018spatial}:
\begin{equation}
  l_{ti}(v_{tj}) =
    \begin{cases}
      0 & \text{if $r_{j}=r_{i}$}\\
      1 & \text{if $r_{j}<r_{i}$}\\
      2 & \text{if $r_{j}>r_{i}$}
    \end{cases}  
\label{eq:Strategy3}
\end{equation}
where $l_{ti}$ presents the label map for each joint $i$ in the neighbor set of the root node $v_{tj}$, $r_{i}$ is the average distance from $cg$ to the root node $v_{tj}$ over each frame and $r_{j}$ is the average distance from $cg$ to the \textit{i}-th joint over each frame across all the training set. Once the labeling of each node in the neighbor set has been set, the convolution operation is performed to produce the output feature maps, as shown in Fig. \ref{fig:spatial_partitioning}.

\subsection{Learnable edge importance weighting}
It is important to note that complex movements can be inferred from a small set of representatives \emph{bright spots} on the joints of the human body \cite{johansson1973visual}. However, not all the joints provide the same quality and quantity of information regarding the movement performed. Therefore, it is intuitive to assign a different level of importance to every joint in the skeleton.\par
In the ST-GCN framework proposed by Yan \textit{et al.}\cite{yan2018spatial}, the authors added a mask M (or M-mask) to each layer of the GCN to express the importance of each joint. The mask applied scales the contribution of each joint of the skeleton according to the learned weights of the spatial graph network. Accordingly, the proposed M-mask considerably improves architecture's performance. Therefore, the M-mask is applied to the ST-GCN network throughout their experiments.

\subsection{Our contribution}
This work proposes an improved set of label mapping methods for the ST-GCN framework by introducing three split processes (full distance split, connection split, and index split) as an alternative approach for the convolution operation. It is based upon the ST-GCN framework proposed by Yan \textit{et al.} \cite{yan2018spatial}. Our results indicate that all of our proposed split strategies outperform the baseline model. Furthermore, the proposed frameworks are more stable during training. Finally, our proposals do not require additional training parameters of the edge importance weighting applied by the ST-GCN model. This proves that our proposal can provide a more suitable solution for real-time applications focused on daily living recognition systems activities for indoor environments.

The contributions are summarized below:
\begin{itemize}
    \item We present an improved set of label mapping methods for the ST-GCN framework by introducing three split processes (full distance split, connection split, and index split) as an alternative approach for the convolution operation. 
    \item Instead of the traditional way of extracting information from the skeleton without considering the relations between the joints, we exploit the relationship between the joints during the action execution to provide valuable and accurate information about the action performed.
    \item We find that an extensive analysis of the inner skeleton joint information by partitioning the skeleton graph in the most number of pieces possible results in more accurate data. 
    \item We propose split strategies that focus on capturing the patterns in the relationship between the skeleton joints by carefully analyzing the partition strategies utilized to perform the movement modeling using the ST-GCN framework.
\end{itemize}.\par

The rest of the paper is structured as follows:
\textbf{Section \ref{section2}} presents state-of-the-art review for previous skeleton graph based action recognition approaches. The details of the proposed skeleton partition strategies are presented in \textbf{Section \ref{proposed_split}}. \textbf{Section \ref{experiment}} discuses the experimental settings we use to obtain the results. The results and discussion are presented in \textbf{Section \ref{results}}. Finally, \textbf{Section \ref{conclusion}} concludes the paper.      

\section{Related Literature}\label{section2}
There has been previous work on activity recognition upon skeleton data. Due to the emergence of low-cost depth cameras, access to skeleton data has become relatively easy \cite{vemulapalli2014human}. Therefore, there has been an increasing interest in using skeleton representations to recognize human activity in general. For the sake of being conscience,  few most recent but relevant works are mentioned. Zhang \textit{et al.} \cite{zhang2019constructing} combined skeleton data with machine learning methods (such as logistic regression) upon dataset benchmarks. They demonstrated that skeleton representations provide better performance in terms of accuracy than other forms of motion representations. In order to model the dependencies between joints and bones, Shi \textit{et al.} \cite{shi2019skeleton} presented a variety of graph networks denominated Directed Acyclic Graph (DAG). Later, Cheng \textit{et al.} \cite{cheng2020skeleton} presented a shift CNN inspired method called Shift- GCN. Their approach aims to reduce the computational complexity of previous ST-GCN-based methods. The results showed the achievement of 10× less computational complexity. However, to the best of our knowledge, there have not been unique partition strategies proposed to enhance the performance of an AR using the ST-GCN model presented in \cite{yan2018spatial}.  

\section{Proposed Split Strategies}\label{proposed_split}
In this section, we present a new set of methods to create the label mapping for the nodes in the neighbor sets of the skeleton graph. The techniques are modifications of the previously proposed spatial configuration partitioning presented in \cite{yan2018spatial}.
As the baseline model, a maximum distance of one node with respect to the root node defines the neighbor sets in the skeleton graph. However, every node in the neighbor set is labeled separately in every strategy presented in this section. Therefore, in every proposed approach, the filter size K = 4. For instance, consider a neighbor set consisting only of the root node with a single adjacent node. For this case, the third and fourth subsets values of the kernel are set by zeros. Each of the split strategies proposed is computed in each frame of a training video sample individually.\par

\begin{figure}[!t]
    \centering
    \includegraphics[width=\linewidth]{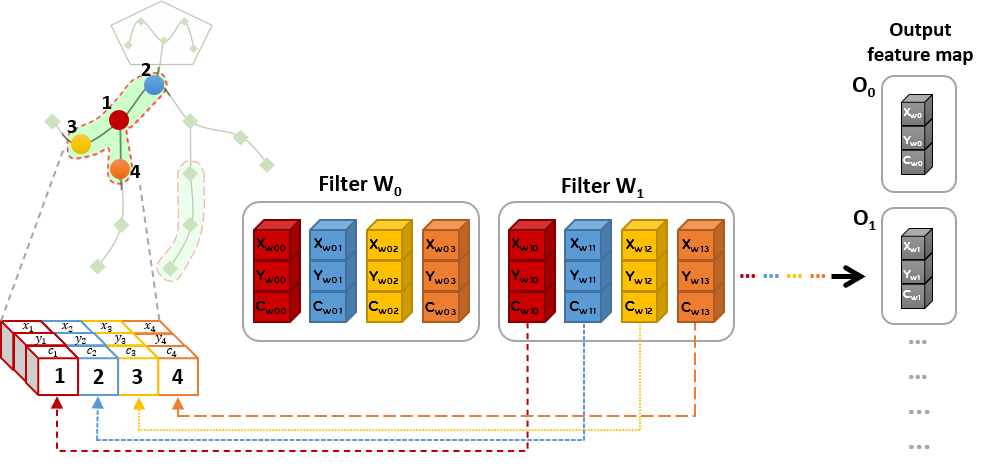}
    \caption{Proposed partition strategy}
    \label{fig:proposed_part}
\end{figure}

Fig. \ref{fig:proposed_part} illustrates our proposed partitioning strategy. As it can be seen, a different label mapping is assigned to each node in the neighbor set. Therefore, a different subset of values of each filter is applied to each joint feature vector. However, the bottleneck is defining each node's order (split criterion) in the neighbor set. We propose three different approaches to address this issue: full distance split, connection split, and index split. These proposals are explained in the following sections.

\begin{figure*}[!t]
    \centering
    \includegraphics[width=0.5\linewidth]{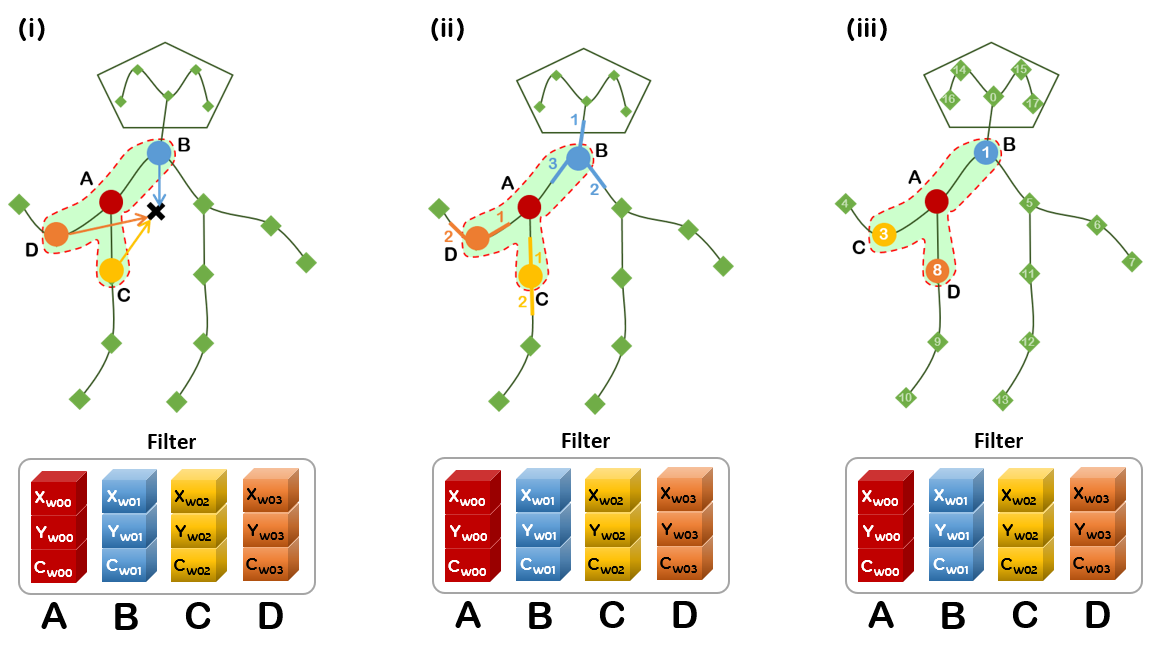}
    \caption{Proposed split methods; (i) Full distance split. (ii) Connection split. (iii) Index split}
    \label{fig:proposed_split}
\end{figure*}

\subsection{Full Distance Split}
In this method, the partitioning for the label mapping is performed according to the distance of each node in the neighbor set with respect to \(cg\). As can be noticed, this solution is similar to the spatial configuration partitioning approach previously explained. However, here we consider the distance of every node in the neighbor set. Thus, this solution is named the \emph{full distance} split method. Therefore, depending on the neighbor set in the skeleton, each kernel can have up to four subsets of values. Fig. \ref{fig:proposed_split}(i) shows that each filter with four subsets of values is applied to the node feature vectors. The order is defined by their relative distances with respect to \(cg\) to create the output feature map. To explain this strategy, we define the set $\mathcal{F}$ as the Euclidean distances of the \textit{i}-th adjacent node $u_{ti}$ (of the root node $u_{tj}$) with respect to \(cg\) sorted in ascending order as:
\begin{equation}
     \mathcal{F}=\{f_{m|m=1,\cdots,N}\}
\end{equation}
where $N$ is the number of adjacent nodes to the root node $u_{tj}$. For instance, $f_{1}$ and $f_{N}$ have the minimum and maximum values in  \(\mathcal{F}\), respectively. In this strategy, the label mapping is given by:
\begin{equation}
  l_{ti}(u_{tj}) =
    \begin{cases}
      0 & \text{if $\|u_{ti}-cg\|_{2}=x_{r}$}\\
      m & \text{if $\|u_{ti}-cg\|_{2}=f_{m}$}\\
    \end{cases}
\end{equation}
where $l_{ti}$ resents the label map for each joint $i$ in the neighbor set of the root node $u_{tj}$, $x_{r}$ is the Euclidean distance from the root node $u_{tj}$ to \(cg\).

\subsection{Connection Split}
In this approach, the number of adjacent joints of each joint (i.e., the joint degree) represents the split criterion in the neighbor set. Thus, the more connections the joint has, the higher priority is assigned to it.
Fig. \ref{fig:proposed_split}(ii) shows that the joint with label A represents the root node, and B is the joint with the highest priority since it has three adjacent joints connected. We observe that both C and D joints have two connections. Hence, the priority for these nodes is set randomly. Once the joint priorities have been set, the convolution operation is performed with a subset of values of each filter for every joint in the neighbor set independently.\par
To define the label mapping in this approach, we first define the neighbor set of a root node $u_{tj}$ and $N$ adjacent nodes as $B(u_{tj})$ \cite{yan2018spatial}, and, we also define the degree matrix of $B(u_{tj})$ as $D$, where $D \in{\mathbb{R}}^{N \times N}$. Therefore, the values at the $d_{ii}$ position of $D$ contain the degree value $d(u_{ti})$ of the each of the adjacent nodes  of the root node $u_{tj}$. Similarly, we define a set $\mathcal{C}$ as the degree values $d(u_{ti})$ of each of the $N$ adjacent nodes of the root node sorted in descending order as follows:
\begin{equation}
    \mathcal{C}=\{c_{m|m=1,\cdots,N}\}
\end{equation}
For instance, $c_{1}$ and $c_{N}$ have the maximum and minimum values of $\mathbcal{C}$, respectively. Finally, the label mapping is thus defined as:
\begin{equation}
  l_{ti}(u_{tj}) =
    \begin{cases}
      0 & \text{if $d(u_{ti})=d_{r}$}\\
      m & \text{if $d(u_{tj})=d_{m}$}\\
    \end{cases}
\end{equation}
where $l_{ti}$ represents the label map for each adjacent joint $i$ to the root node $u_{tj}$ in the neighbor set, and $d_{r}$  is the degree corresponding the root node $u_{tj}$

\subsection{Index Split}
\begin{figure}[h]
    \centering
    \includegraphics[width=0.2\textwidth]{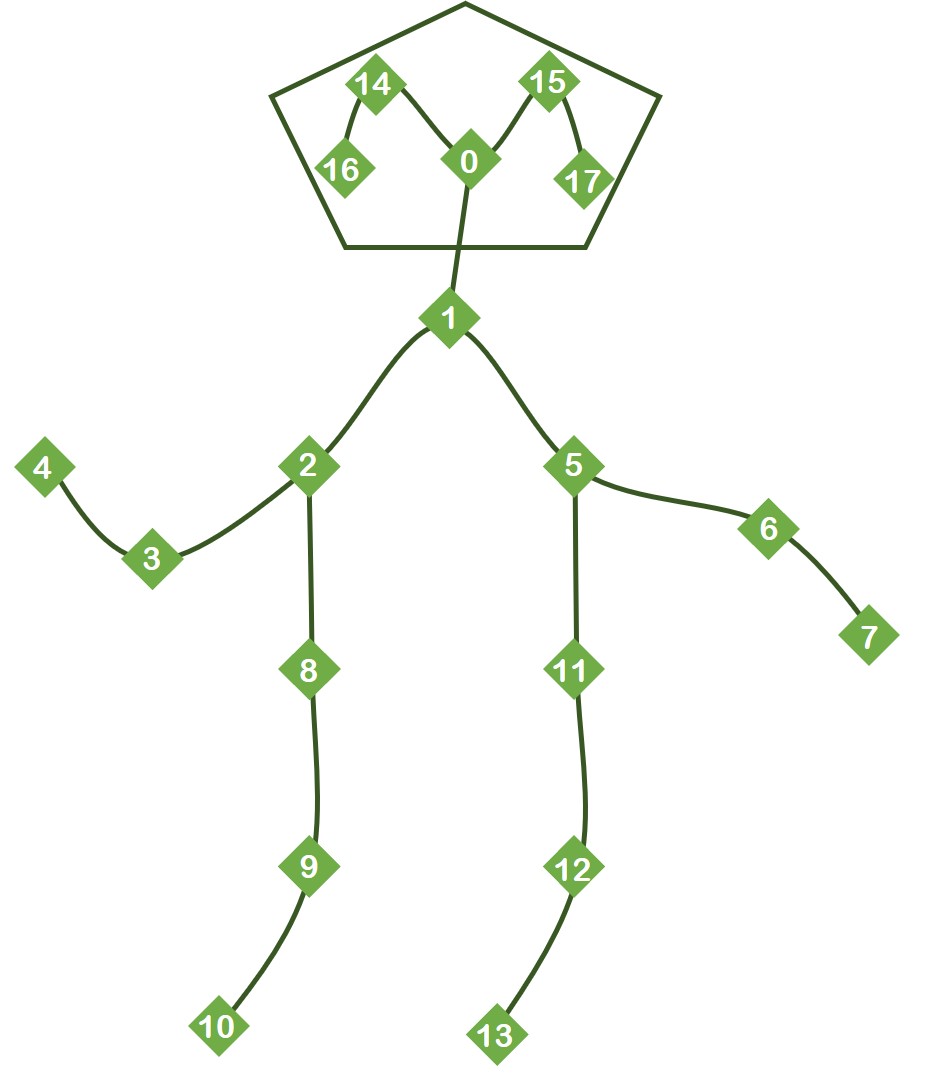}
    \caption{Open-Pose output keypoints.}
    \label{fig:open_pose_keypoints}
\end{figure}

The skeleton data utilized for our study is gathered using the Open-Pose \cite{cao2019openpose} library. According to the library documentation, the output file with the skeleton information consists of critical/key points. The output skeleton provided by the Open Pose toolbox is shown in Fig. \ref{fig:open_pose_keypoints}.\par

In this approach, the value of the index of each key point defines the priority criterion of the neighbor set. An illustrative example is shown in Fig. \ref{fig:proposed_split}(iii).  For instance, joint B is assigned with the highest priority since it has a key-point index value of 1, and C is the joint with the second priority since it has a keypoint index value of 3. Finally, D is the joint with the least priority since it has a keypoint index value of 8.
Therefore, we define the set $\mathbcal{P}$ as the indexes of the key points $ind(u_{ti})$ of the \textit{i}-th adjacent nodes $u_{ti}$ (of the root node $u_{tj} $) sorted in ascending order as:
\begin{equation}
    \mathcal{P}=\{p_{m|m=1,\cdots,N}\}
\end{equation}
where $N$ is the number of adjacent nodes to the root node $u_{tj}$. The label mapping is therefore defined as:
\begin{equation}
  l_{ti}(u_{tj}) =
    \begin{cases}
      0 & \text{if $ind(u_{ti})=in_{r}$}\\
      m & \text{if $ind(u_{tj})=p_{m}$}\\
    \end{cases}
\end{equation}
where $l_{ti}$ represents the label map for each joint $i$ in the neighbor set of the root node $u_{tj}$ and $in_{r}$ is the index of the keypoint corresponding to the root node $u_{tj}$.
\section{Experiments}\label{experiment}
\subsection{Datasets}
To evaluate the performance of our proposed partitioning schemes, we train our models on two benchmark datasets: the NTU RGB+D \cite{shahroudy2016ntu} and the Kinetics \cite{kay2017kinetics} dataset. These two datasets were considered in order to provide a valid comparison with the original ST-GCN framework.
\subsubsection*{NTU-RGB+D}
Up to date, the NTU-RGB+D is known to be the most extensive dataset with 3D joints annotations for human action recognition tasks \cite{yan2018spatial}. The samples have been recorded using the Microsoft Kinect V2 camera. In order to take the most advantage of the chosen camera device, each action sample consists of a depth map modality, 3D joint information, RGB frames, and IR sequences. The information provided by this dataset consists of the tri-dimensional location of the 25 main joints of the human body.\par
In their study, Shahroudy \emph{et al.} \cite{shahroudy2016ntu} proposed two evaluation criteria for the NTU-RGB+D dataset: the Cross-Subject (\emph{X-sub}) and the Cross-View (\emph{X-view}) evaluations. In the first approach, the train/test split for evaluation was based upon groups of subjects performing the action; the data corresponding to 20 participants is used for training and the remaining samples for testing. On the other hand, the Cross-View evaluation approach considers the camera view as criteria for the train/test split; the data collected by the camera 1 is used for testing and the data collected by the other two cameras is used for training.\par
The NTU-RGB+D dataset provides a total of 56,880 action clips performing 60 different actions classified into three major groups: daily actions, health-related actions, and mutual actions. Forty participants performed the test action samples. Each sample has been captured with 3 different cameras simultaneously located at the same height but different angles. Later, this dataset was extended twice its size by adding 60 more classes and another 57,600 video samples \cite{shahroudy2016ntu}. This extended version is called NTU RGB+D 120 (120-class NTU RGB+D dataset). By considering the 3D skeletons modality of the NTU-RGB+D dataset only, the storage was reduced from 136 GB to 5.8 GB. Therefore, the computational speed is reduced considerably.
\subsubsection*{Kinetics}
While the NTU-RGB+D dataset is widely known to be the largest in-house captured action recognition dataset, the Deepmind Kinetics human action dataset is the largest set with unconstrained action recognition samples. 
The 306,245 videos provided by the Kinetics dataset are obtained from YouTube.  Each video sample is supplied with no previous editing to ensure good variable resolution and frame rate for action modeling and is classified into 400 different action classes.\par
Due to the vast quantity of classes, one video sample can be classified into more than one cluster. For instance, a video sample with a person texting while driving a car can be classified with the “texting” label or the “driving a car” label. Therefore, the authors in \cite{kay2017kinetics} suggest to consider a top-5 performance evaluation rather than a top-1 approach. Meaning that, a labelled sample is considered a true positive if its ground truth label appears within the 5 classes with the highest scores predicted by the model (top-5); contrary to considering only the predicted class with the highest score (top-1).\par
The Kinetics dataset provides the raw RGB format videos. Therefore, it requires the skeleton information to be extracted from the sample videos. Accordingly, we use the dataset that contains the Kinetics-skeleton information provided by Yan \textit{et al.} \cite{yan2018spatial} for our experiments.

\subsection{Model Implementation}
The experiment process comprises of three stages: Data Splitting, ST-GCN model setup, and Model Training. These stages are explined as follows:\par

\subsubsection*{Data Splitting}
The datasets is divided into two subsets: the training and the validation sets. In our experiments, we consider a 3:1 relation for training and validation split, respectively. 
\subsubsection*{ST-GCN Model Setup}
The ST-GCN model uses a baseline architecture. It consists of a stack of 9 layers that are divided into 3-layer blocks stacked together. Each layer block consists of 3 layers each. The layers of the first block have 64 output channels each. The second and third blocks have 128 and 256 output channels, respectively. Finally, the 256 feature vector output by the last layer is fed into a Softmax classifier to predict the performed action \cite{yan2018spatial}.

\subsubsection*{Model Training}
The ST-GCN model is implemented on the PyTorch framework for deep learning modelling \cite{paszke2017automatic}. The models are trained using stochastic gradient descent (SGD) with learning rate decay as an optimization algorithm. The initial learning rate is 0.1. The number of epochs and decay schedule for training varies depending on the dataset used. For the NTU-RGB+D dataset, we train the models for 80 epochs, and the learning rate decays by a factor of 0.1 on the 10\textit{th} and the 50\textit{th} epochs. On the other hand, for the Kinetics dataset, we train the models for 50 epochs, and the learning rate decays by a factor of 0.1 every 10\textit{th} epochs.
Similarly, the batch size also varies according to the dataset utilized; for the NTU-RGB+D dataset, the batch sizes for training and testing used were 32 and 64, respectively; on the other hand, for the Kinetics dataset, the batch sizes for training and testing used were 128 and 256, respectively. To avoid overfitting, a weight decay value of 0.0001 has been considered. Additionally, a dropout value of 0.5 has been set for the NTU-RGB+D dataset experiments.
To provide a valid comparison with the baseline model, an M-mask implementation is considered in the experiments presented in this study.
\section{Experimental Results and Discussion}\label{results}
This section discusses the performance of our proposals against the benchmark ST-GCN models based on \cite{yan2018spatial} using the spatial configuration partition approach. This strategy provides the best performance in terms of accuracy in \cite{yan2018spatial}. Therefore, it has been chosen as a baseline to prove the effectivenes of the partition strategies introduced in this study.  
\subsection{Results Evaluation on NTU-RGB+D}
Note that we aim to recognize ADL in an indoor environment. Therefore, the NTU-RGB+D dataset serves as a more accurate reference than the Kinetics dataset since it was recorded using the same conditions. Hence, we focus on the results obtained with this dataset. We use the 3D joint information provided in \cite{shahroudy2016ntu} in our experiments. The Table \ref{tab:experimental_NTU-RGB_D} shows the performance comparisons of our proposals and the state-of-the-art ST-GCN framework. It can be observed that all of our partition strategies outperform the spatial configuration strategy of the ST-GCN. For the X-sub benchmark, the connection split achieves the highest performance of 82.6\% accuracy, more than 1\% higher than the ST-GCN performance. On the other hand, the index split outperforms the rest of the strategies with 90.5\% accuracy on the X-view benchmark, more than 2\% higher than the ST-GCN performance.

\begin{table}[!tb]
\caption{NTU-RGB+D Performance}
\label{tab:experimental_NTU-RGB_D}
\centering
\begin{tabular}{|c|l|l|l|}
\hline
\multicolumn{2}{|c|}{Method}                                      & X-sub           & X-view          \\ \hline
\multicolumn{1}{|l|}{ST-GCN} & Spatial configuration partitioning & 81.5\%          & 81.5\%          \\ \hline
\multirow{3}{4em}{Ours}      & Full distance split                & 81.6\%          & 89.3\%          \\ \cline{2-4} 
                             & Connection split                   & \textbf{82.6\%} & 89.6\%          \\ \cline{2-4} 
                             & Index split                        & 81.7\%          & \textbf{90.5\%} \\ \hline
\end{tabular}
\end{table}

\begin{figure*}[!t]
\begin{subfigure}{.5\textwidth}
  \centering
  \includegraphics[width=0.8\linewidth]{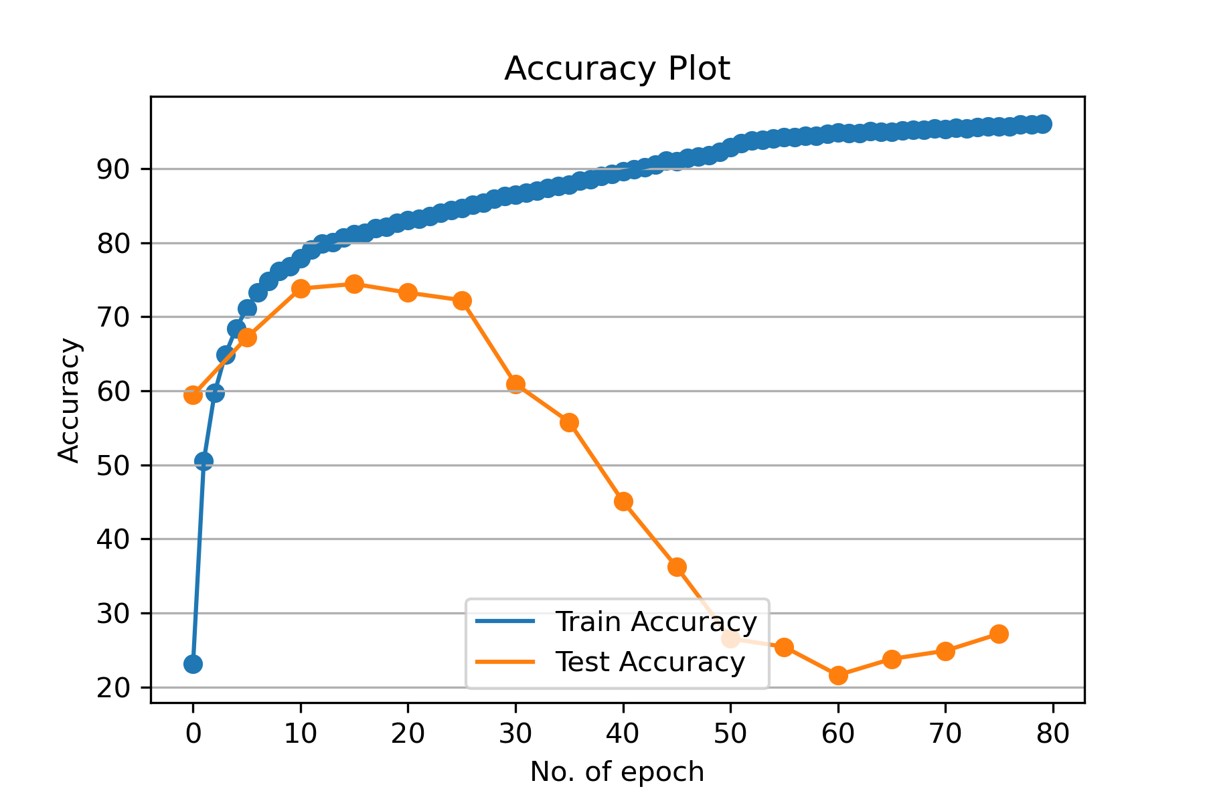}  
  \caption{Spatial C.P X-sub training scores}
  \label{fig:spatial_x1}
\end{subfigure}
\begin{subfigure}{.5\textwidth}
  \centering
  \includegraphics[width=0.8\linewidth]{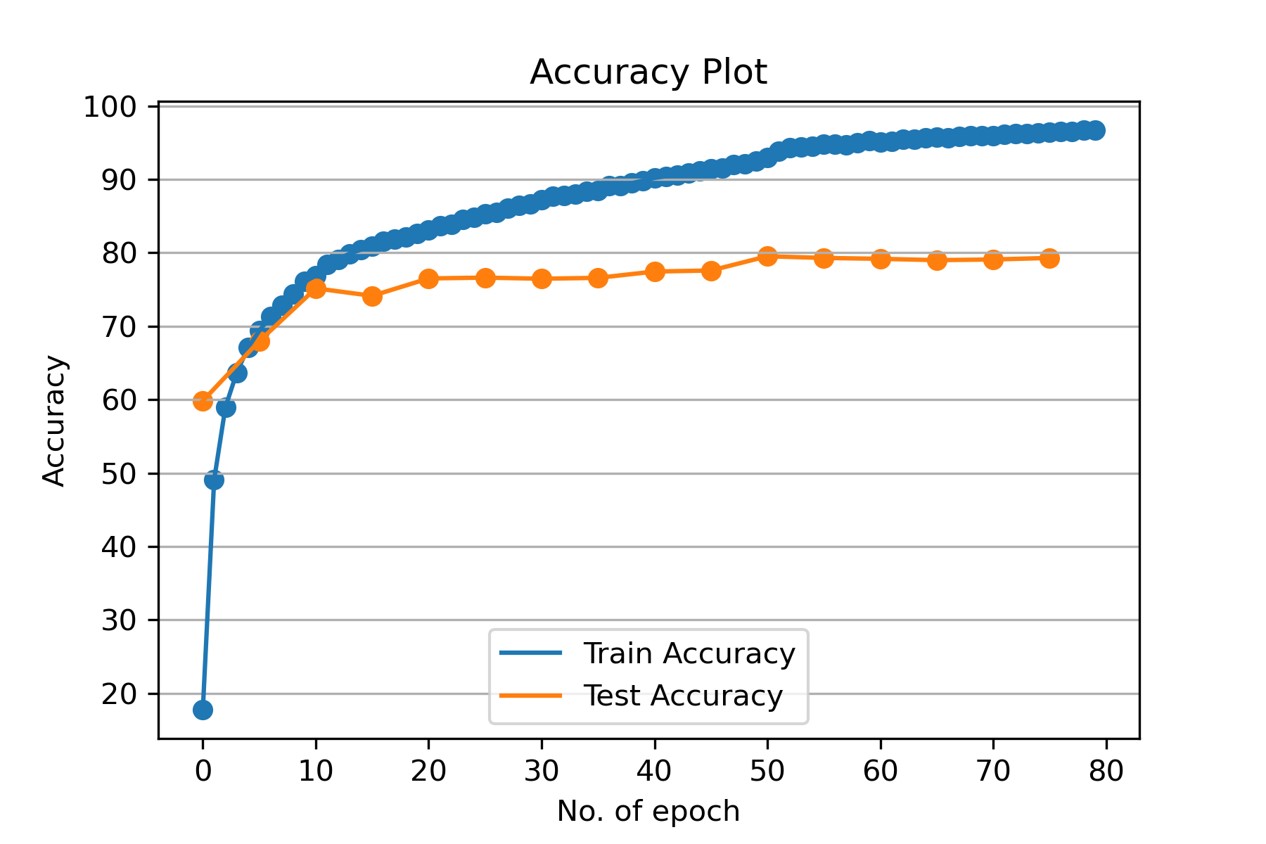}  
  \caption{Connection split X-sub training scores}
  \label{fig:connection_x1}
\end{subfigure}

\begin{subfigure}{.5\textwidth}
  \centering
  \includegraphics[width=0.8\linewidth]{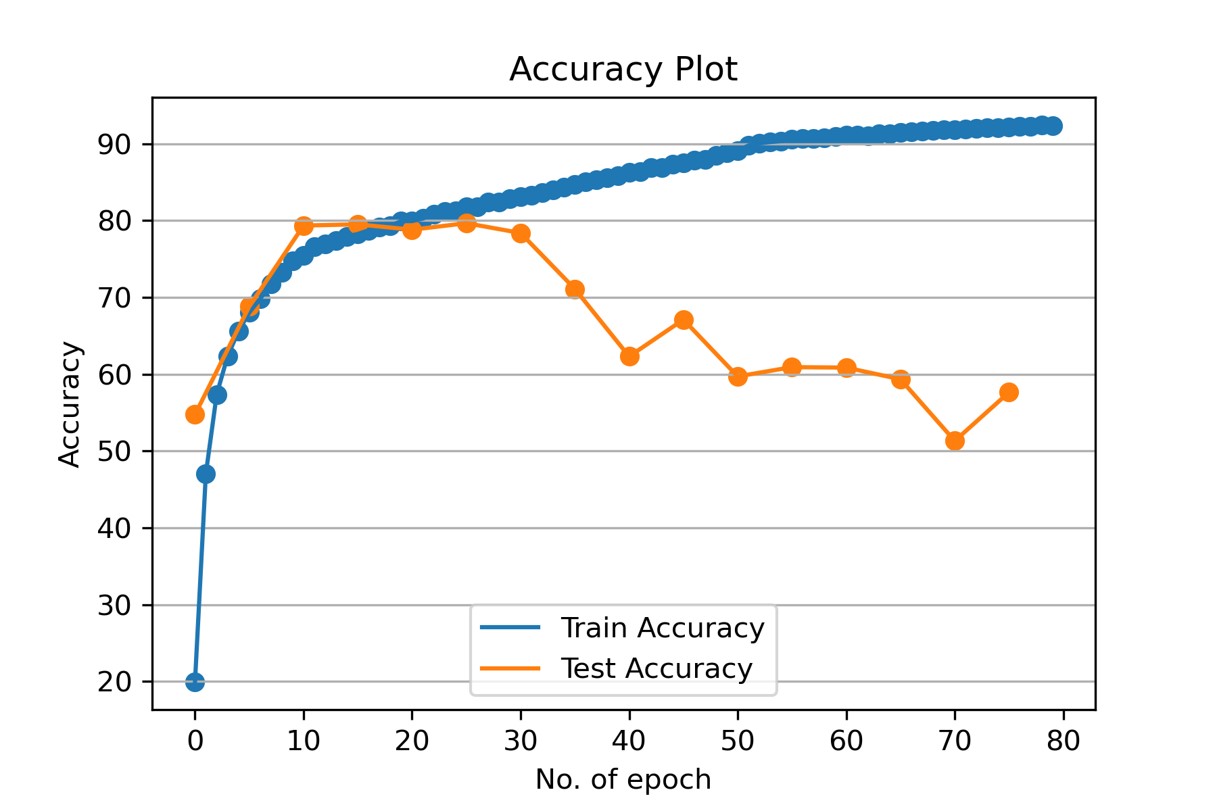}  
  \caption{Spatial C.P X-view training scores}
  \label{fig:spatial_x2}
\end{subfigure}
\begin{subfigure}{.5\textwidth}
  \centering
  \includegraphics[width=0.8\linewidth]{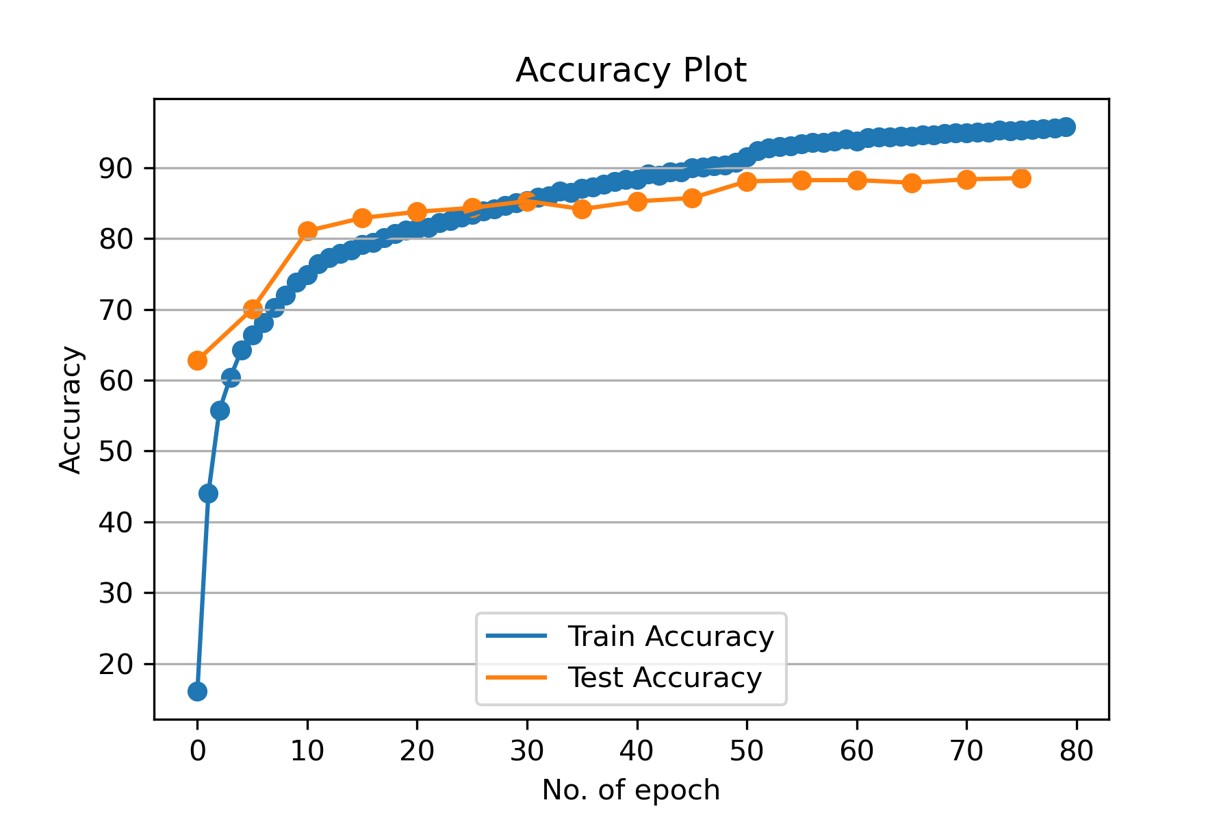}  
  \caption{Connection split X-view training scores}
  \label{fig:connection_x2}
\end{subfigure}
\caption{Training scores}
\label{fig:accuracy_score}
\end{figure*}

Figs \ref{fig:spatial_x1}, \ref{fig:connection_x1}, \ref{fig:spatial_x2} and \ref{fig:connection_x2} show the training behavior of the models using the spatial configuration partitioning of the ST-GCN framework and the proposed connection split on both X-sub and X-view benchmarks without the M-mask implementation. The blue and orange plots show the performance of the models using the training and the validation sets, respectively. The training score plots show that the learning performance of the proposed connection split stabilizes while increasing over time compared with the ST-GCN outcome. Our proposals provide a considerable advantage over the benchmark framework because it demonstrates that the M-mask is not required to yield satisfactory performance. The omission of the M-mask results in a reduction of computational complexity. Hence, our proposal can provide a more suitable solution for real-time applications. Moreover, given the performance superiority on accuracy and time consumption, our proposed method offers a practical solution an ADL recognition system.
\subsection{Performance on the Kinetics Dataset}
The recognition performance has been evaluated using the top-1 and top-5 criterion using the Kinetics dataset. We validate the performance of our proposed methods with the ST-GCN framework, as shown in Table \ref{tab:experimental_Kinetics}.

\begin{table}[!t]
\caption{Performance on Kinetics Dataset}
\label{tab:experimental_Kinetics}
\centering
\begin{tabular}{|c|l|l|l|}
\hline
\multicolumn{2}{|c|}{Method}                                      & Top–1           & Top–5          \\ \hline
\multicolumn{1}{|l|}{ST-GCN} & Spatial configuration partitioning & 30.7\%          & 52.8\%          \\ \hline
\multirow{3}{4em}{Ours}   &Full distance split & \textbf{31.7\%}  & \textbf{54.5\%} \\ \cline{2-4} 
                        & Connection split  & 30.7\% & 53.3\%   \\ \cline{2-4} 
                        & Index split   & 31.5\% & 54.1\% \\ \hline
\end{tabular}
\end{table}
As the results indicate, all of our partition strategies outperform the spatial configuration strategy of the ST-GCN using the top-5 criteria. We observe that 54.5\% accuracy is achieved using the full distance split approach, which is 2\% higher than the performance obtained with the baseline model. On the other hand, by using the top-1 evaluation criteria, our proposal achieves the same performance as the ST-GCN model. Similarly, using this evaluation basis, the highest performance achieved is a 31.7\% accuracy using the full distance split approach resulting in a 1\% margin higher than the result obtained with the ST-GCN model.
Therefore, we can conclude that the performance metrics presented in Table \ref{tab:experimental_Kinetics} validates the superiority of the full distance split method proposed on the Kinetics dataset.

\section{Conclusion}\label{conclusion}
In this work, we propose an improved set of label mapping methods for the ST-GCN framework (full distance split, connection split, and index split) as an alternative approach for the convolution operation. Our results indicate that all of our split processes outperform the previous partitioning strategies for the ST-GCN framework. Moreover, they demonstrate to be more stable during training without using the additional training parameter of the edge importance weighting applied by the baseline model. Therefore, the results obtained with our current split proposals can provide a more suitable solution for real-time applications focused on activities of daily living recognition systems for indoor environments than the baseline strategies for the ST-GCN framework.\par
A significant computational effort is involved in using heterogeneous methods to calculate the distances between the joints and the $cg$  for each frame in the video sample for full distance split and spatial configuration partitioning. It will be computationally less demanding to use a homogeneous technique to calculate the distance between the joints and the $cg$ for both splitting strategies. Furthermore, while our current methodology considers greater distances from the root node to perform the skeleton partitioning, additional flexibility can be made by increasing the amount joints per neighbor set. This may give room to cover larger body sections (such as limbs), making it possible to find more complex relationships between the joints during the execution of the actions.

\bibliographystyle{IEEEtran}
\bibliography{ref}

\end{document}